\documentclass[review]{elsarticle}

\label{sec:packages}
\usepackage[rgb,x11names]{xcolor}
\usepackage{hyperref}
\usepackage{subfig}
\usepackage[draft]{todonotes}
\usepackage{algorithm}
\usepackage{algorithmicx}
\usepackage{algpseudocode}
\usepackage{siunitx}
\usepackage{newtxmath}
\usepackage{tabularx}
\usepackage{booktabs}
\usepackage{array}
\usepackage{multirow}

\def\ourmodel{MAST}
\newcommand{\myPara}[1]{\noindent\textbf{#1}}
\newcommand{\secref}[1]{$\S$ \ref{#1}}
\newcommand{\figref}[1]{Fig. \ref{#1}}
\newcommand{\mytodo}[1]{\todo{TODO}}

\definecolor{mygray}{gray}{.92}
\def\ie{\emph{i.e.}}

\def\etc{\emph{etc}}
\def\etal{{\em et al. }}

\journal{Pattern Recognition}

\begin{document}

\begin{frontmatter}



    \title{\textbf{MAST: Video Polyp Segmentation with a Mixture-Attention Siamese Transformer}}

    \author[1]{Geng Chen}
    \author[1]{Junqing Yang}

    \author[1]{Xiaozhou Pu}

    \author[2]{Ge-Peng Ji}

    \author[3,4]{Huan Xiong}

    \author[1]{Yongsheng Pan}

    \author[1]{Hengfei Cui}

    \author[1]{Yong Xia\corref{cor}}
    \ead{yxia@nwpu.edu.cn}
    \cortext[cor]{Corresponding author}
    \cortext[equal]{G. Chen, J. Yang, and X. Pu contribute equally.}

    \address[1]{National Engineering Laboratory for Integrated Aero-Space-Ground-Ocean Big Data Application Technology, School of Computer Science and Engineering, Northwestern Polytechnical University, Xi’an, China.}
    \address[2]{School of Computing, Australian National University, Canberra, Australia.}
    \address[3]{Mohamed bin Zayed University of Artificial Intelligence, UAE.}
    \address[4]{Harbin Institute of Technology, China.}

    \begin{abstract}
        Accurate segmentation of polyps from colonoscopy videos is of great significance to polyp treatment and early prevention of colorectal cancer. However, it is challenging due to the difficulties associated with modelling long-range spatio-temporal relationships within a colonoscopy video. In this paper, we address this challenging task with a novel \textbf{M}ixture-\textbf{A}ttention \textbf{S}iamese \textbf{T}ransformer (\textbf{\ourmodel}), which explicitly models the long-range spatio-temporal relationships with a mixture-attention mechanism for accurate polyp segmentation. Specifically, we first construct a Siamese transformer architecture to jointly encode paired video frames for their feature representations. We then design a mixture-attention module to exploit the intra-frame and inter-frame correlations, enhancing the features with rich spatio-temporal relationships. Finally, the enhanced features are fed to two parallel decoders for predicting the segmentation maps. To the best of our knowledge, our \ourmodel~is the first transformer model dedicated to video polyp segmentation. Extensive experiments on the large-scale SUN-SEG benchmark demonstrate the superior performance of \ourmodel{} in comparison with the cutting-edge competitors. Our code is publicly available at \url{https://github.com/Junqing-Yang/MAST}.
    \end{abstract}


\end{frontmatter}

\section{Introduction}\label{Introduction}
Colorectal cancer (CRC) is a major cause of cancer-related deaths globally \cite{biller2021diagnosis}.
As a widely used screening test for CRC, colonoscopy is used by physicians to examine polyps, which may develop into cancer if left untreated \cite{arnold2017global}.
Manual examination is highly dependent on physician experience and judgment, with high rates of missed and misdiagnosed polyps.
According to existing studies \cite{ahn2012miss}, more than 20\% of colon polyps are missed/misdiagnosed during endoscopy, underscoring the need for better detection methods.
Automatic polyp segmentation can significantly reduce physicians' workload and improve polyp detection accuracy, making it a crucial tool for CRC screening and prevention.

However, automatic polyp segmentation is a challenging task \cite{ji2023sam} since polyps are highly variable in appearance (e.g., shape, size, color) and the colonoscopy images suffers from quality issues (e.g., low contrast, noise, artifacts, specular reflections).
Furthermore, polyps can be easily misidentified with other enteric tissues, such as blood vessels and feces.
To address these challenges, significant efforts \cite{ronneberger2015u,brandao2017fully,zhou2019unetplusplus,fan2020pranet,wei2021shallow,mahmud2021polypsegnet,yeung2021focus,zhang2022lesion} have been made to identify polyps using deep learning techniques, which show excellent performance in image segmentation.
Most of exciting methods focus on segmenting polyps from colonoscopy images rather than videos.
For instance, Fan \etal \cite{fan2020pranet} proposed PraNet, which used the reverse attention module to mine and model relationships between regions and boundaries.
Wei \etal \cite{wei2021shallow} focused on the shallow features of the image and used shallow attention module to eliminate the noise and fully explore the shallow information of the image.
However, the image-based methods overlook the vital clues in the temporal context of the video, limiting the accuracy of segmentation.

Instead of relying on images, methods have been proposed to use colonoscopy videos fully.
These methods are categorized as video polyp segmentation (VPS), where convolutional neural networks (CNNs) have been widely employed \cite{puyal2020endoscopic,ji2021progressively,ji2022video,wu2021multi,wu2021precise,li2022tccnet,zhao2022semi}.
{For instance, Puyal \etal \cite{puyal2020endoscopic} proposed a hybrid VPS framework, where a 2D network acts as the backbone for extracting spatial features and a 3D network ensures temporal consistency.}
Ji \etal \cite{ji2022video} comprehensively introduced the work related to video polyp segmentation in deep learning and the proposed model, PNS+, is the first to introduce a high-quality fine-grained annotated VPS dataset named SUN-SEG \cite{misawa2021development}.

Existing works have made progress in the VPS task, but several challenges remain.
One key challenge is how to model temporal relationships among consecutive frames, which is difficult due to variations in polyps over time.
Another challenge is that CNNs may not fully capture long-range relationships, which are crucial for segmenting polyps with large shape variations or low boundary contrast.
Unlike CNNs, transformers show particularly good performance in modelling long-range relationships, and they have seldomly been investigated for the VPS task.

To this end, we propose a \textbf{M}ixture-\textbf{A}ttention \textbf{S}iamese \textbf{T}ransformer (\textbf{\ourmodel}) for accurate VPS from colonoscopy videos, as shown in \figref{fig:model_structure}.
\ourmodel{} proposes a mixture-attention mechanism to model the spatio-temporal relationships, designs a Siamese architecture for learning from video frames jointly and employs a transformer to learn the long-range relationships inner frames.
Specifically, we first use a pair of transformers with shared weights as a Siamese backbone to extract features.
The resulting features are then fed to our mixture-attention module, which jointly integrates inter-frame mutual-attention and intra-frame self-attention into a unified framework to exploit the long-range spatio-temporal relationships of two frames for improved feature representation learning.
Finally, the refined features are passed through two decoders for predicting the polyp segmentation maps.
Our \ourmodel{} overcomes the challenges associated with accurate VPS with a novel mixture-attention mechanism and a Siamese transformer architecture.
Extensive experiments on the mainstream SUN-SEG benchmark demonstrate the superior performance of \ourmodel{} over other cutting-edge VPS models.
The main contributions of our work can be summarized as follows:
\begin{itemize}
    \item[$\bullet$] We design a Siamese transformer to jointly encode paired video frames, providing rich features for accurate VPS.
    \item[$\bullet$] We propose a mixture-attention module to simultaneously mine inter- and intra-frame long-range relationships, enhancing the features to promote the accuracy of VPS.
    \item[$\bullet$] Our \ourmodel{} significantly promotes the spatio-temporal learning ability, setting the new state-of-the-art on the challenging SUN-SEG benchmark.
\end{itemize}

Our paper is organized as follows:
Section \secref{sec:related} introduces the relevant works, including polyp segmentation, visual transformer, and attention mechanism;
Section \secref{sec:method} describes the architecture of our \ourmodel{} model along with the Siamese transformer, mixture-attention module, parallel decoders, and loss function;
Section \secref{sec:Experiments} presents the experimental results and ablation study;
Finally, Section \secref{sec:Conclusion} summarizes this work.

\section{Related Work}\label{sec:related}
This section reviews the relevant works in video polyp segmentation (see Section \secref{sec:Polyp}), transformer in vision (see Section \secref{sec:Transformer}), and attention mechanisms (see Section \secref{sec:attention}).

\subsection{Polyp Segmentation}\label{sec:Polyp}
Early polyp segmentation methods rely on hand-crafted features, such as texture and colour \cite{tajbakhsh2015automated}, intensity distribution \cite{jerebko2003polyp}, geometric features \cite{mamonov2014automated}, \etc.
However, due to the large appearance variation of polyps and the high similarity between polyps and surrounding normal tissues, traditional methods have a high rate of missed diagnosis and misdiagnosis.

Deep learning has been employed for more accurate image polyp segmentation with rich features automatically learned by the networks \cite{brandao2017fully,ronneberger2015u,zhou2019unetplusplus,yeung2021focus,mahmud2021polypsegnet,lin2022bsca}.
For instance, the works in \cite{brandao2017fully} apply a fully convolutional network to identify and segment polyps from colonoscopy images.
In recent years, U-shape networks \cite{ronneberger2015u,zhou2019unetplusplus} have been widely adopted for poly segmentation due to their excellent performance in medical image analysis tasks.
Focus U-Net \cite{yeung2021focus} combines U-Net and attention components into a focus gate to {control the degree of background suppression}.
PolypSegNet \cite{mahmud2021polypsegnet} uses a deep fusion jump module instead of the original jump connection.
Apart from these, there are also non-U-shape networks proposed for poly segmentation.
Besides, there are many other methods for image polyp segmentation work \cite{fan2020pranet,Tomar2022TGANet,wu2021collaborative}.
PraNet \cite{fan2020pranet} uses the inverse attention module to mine and model relationships between regional and boundary cues.
Typical methods include PraNet \cite{fan2020pranet} and the method in \cite{wu2021collaborative}, where an adversarial training framework is proposed and employed to deal with the diversity of polyp location and shape through focusing and dispersion extraction.

In the early years, limited by datasets and networks, most polyp segmentation works were based on images.
Instead of relying on images, efforts have been dedicated to VPS that directly segment polyps from colonoscopy videos.
Hybrid 2/3D CNN framework \cite{puyal2020endoscopic} is used to aggregate spatio-temporal correlation and obtain better segmentation results.
PNS+ \cite{ji2022video} is the first study to comprehensively introduce the work related to video polyp segmentation in deep learning and the first to introduce a high-quality fine-grained annotated VPS dataset named SUN-SEG \cite{misawa2021development}.
At the same time, a global encoder and a local encoder are designed in PNS+ to extract the long-term and short-term feature representation, respectively, and introduce a self-attention block to update the receptive field dynamically. PNS+ achieves the most advanced performance to date.
Other related studies, such as STFT \cite{wu2021multi} and SCR-Net \cite{wu2021precise}, have also explored the task of VPS and achieved promising results.

However, none of the aforementioned VPS methods considers a transformer in their work and lacks explicit modelling of spatio-temporal relationships within a colonoscopy video. These issues are resolved by our \ourmodel{}, therefore leading to cutting-edge performance.

\subsection{Transformer in Vision}\label{sec:Transformer}
Inspired by the success of transformers in natural language processing, many studies are exploring its application to computer vision.
Since then, the transformer has made its mark in the computer vision tasks, such as image classification, object detection, semantic segmentation, image generation, video understanding \etc.
ViT \cite{dosovitskiy2021vit} divides images into fixed-sized patches, sends patch embedding vector to transformer encoder after coding, and then uses MLP to perform image classification. However, ViT is only suitable for simple classification tasks, and its performance is poor in pixel-wise dense prediction scenarios. To handle these advanced visual tasks.
Pyramid Vision Transformer (PVT) \cite{wang2021pvt}, which uses fine-grained image blocks as input, is proposed to solve the downstream semantic segmentation task. It introduces a progressive shrinking pyramid to reduce the transformer sequence length and significantly reduce the computational cost with the deepening of the network.
PVTv2 \cite{wang2022pvt} improves the component linear complexity attention layer, overlapping patch embedding on the original PVT and convolutional feed-forward network.
The computational complexity of PVTv2 is reduced to linearity, resulting in significant improvements to basic visual tasks such as classification, detection, and segmentation. Interested readers can refer to \cite{Han2023survey} for a comprehensive literature review of transformers in vision.

\subsection{Attention Mechanisms.}\label{sec:attention}
Inspired by human vision, Mnih \etal \cite{Mnih2014Recurrent} first applied the Attention mechanism in computer vision for the image classification task and achieved excellent results.
Then Attention mechanism is widely used in various tasks \cite{xu2015show,zhang2021learning} based on RNN/CNN and other neural network models \cite{Wang2020ISO}.
Moreover, all kinds of attention mechanics are coming, including spatial attention \cite{Woo2018CBAM,yue2023attention}, channel attention \cite{Hu2018Squeeze}, and self-attention \cite{Cao2019GCNet,Lu2022Zero}, to name a few.
The self-attention mechanism is a variant of the attention mechanism, which reduces the dependence on external information and is better at capturing the internal relevance of data or features.
The self-attention mechanism's application in CV mainly solves the long-range dependence problem by calculating the mutual influence between patches.

Most attention is focused on each modality and context individually, so co-attention is proposed \cite{Lu2016Hierarchical}.
Furthermore, it is successfully applied to crossover or cross-modal tasks \cite{Wu2018are,Nguyen2018Improved}.
The co-attention mechanism allows the network to project different modes into a common feature space and effectively mine the potential associations between them.
Unlike existing works, we propose a mixture-attention mechanism that integrates cross-frame co-attention and intra-frame self-attention into a unified framework, allowing full capturing of the long-range relationships in the colonoscopy videos.

\begin{figure}[t]
	\centering
	\includegraphics[trim={0cm 1cm 0cm 1cm},clip,width=\linewidth]{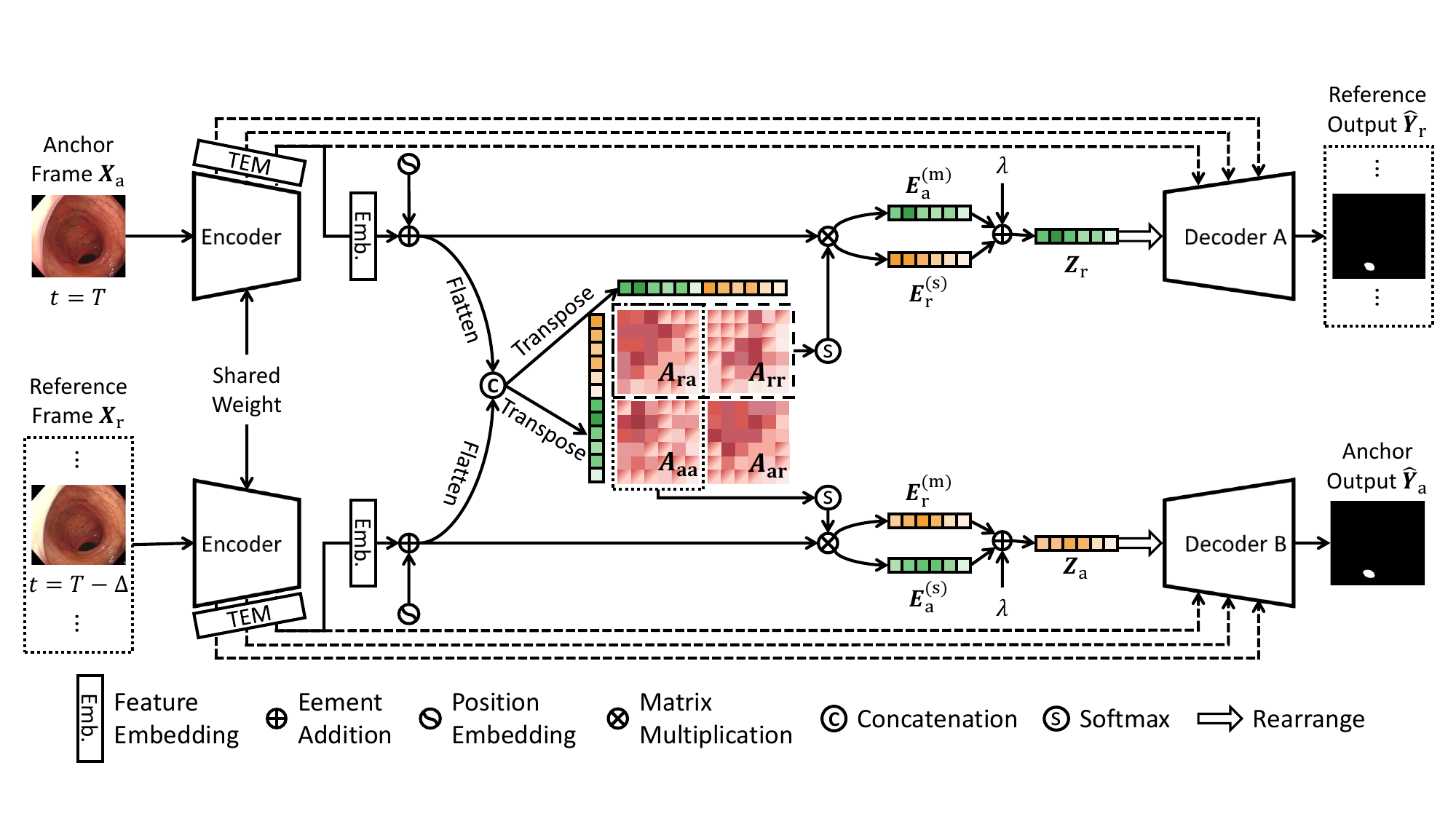}
	\caption{
		\textbf{Overview of MAST:} Using an anchor frame at $t=T$ and a reference frame at $t=T-\Delta$ as input, MAST employs a Siamese transformer for feature extraction and a Mixture-Attention module to compute the attention matrix to enhance the features. The enhanced features are then fed into two parallel decoders for segmentation map prediction.
	}
	\label{fig:model_structure}
\end{figure}
\section{Method}\label{sec:method}
In this section, we provide detailed descriptions for the key modules of our \ourmodel{}, including Siamese transformer (see Section \secref{sec:SiameseTransformer}), Mixture-Attention module (see Section \secref{sec:MixtureAttention}), the decoders (see Section \secref{sec:ParallelDecoders}) and the loss function (see Section \secref{sec:LossFunction}).

\subsection{Siamese Transformer}\label{sec:SiameseTransformer}
We develop a Siamese transformer to jointly extract rich features from paired video frames with a high efficiency.
As shown in the far left of \figref{fig:model_structure}, we sample pairwise frames from a given colonoscopy video, including an anchor frame $\boldsymbol{X}_{\text{a}}$ at time $t=T$ and a reference frame $\boldsymbol{X}_{\text{r}}$ at time $t=T-\Delta$.
In general, the Siamese transformer consists of three major components, including batch formation (\ie, $\operatorname{BatchForm}(\cdot)$), transformer (\ie, $\operatorname{Transformer}(\cdot)$), and batch split (\ie, $\operatorname{BatchSplit}(\cdot)$).
Three components are orgnalized in a cascaded manner detailed as follows.
First, we create a batch using two paired fames with:
\begin{equation}  \label{equation_BatchForm}
    \boldsymbol{\hat{X}} = \operatorname{BatchForm}(\boldsymbol{X}_{\text{a}}, \boldsymbol{X}_{\text{a}}).
\end{equation}

$\boldsymbol{\hat{X}}$ is then fed into a transformer \cite{wang2022pvt} to learn the multi-level side-out features $\{\boldsymbol{\hat{F}}^{(i)}\}_{i}$, \ie,
\begin{equation}  \label{equation_SiamTrans}
    \{\boldsymbol{\hat{F}}^{(i)}\}_{i} = \operatorname{Transformer}(\boldsymbol{\hat{X}}).
\end{equation}

For clarity, we omit the superscript $i$ and use $\boldsymbol{\hat{F}}$ to denote the last layer side-out feature map. $\boldsymbol{\hat{F}}$ is then fed to a batch split component:
\begin{equation}  \label{equation_BatchSplit}
    [\boldsymbol{\hat{F}}_{\text{a}}, \boldsymbol{\hat{F}}_{\text{r}} ] = \operatorname{BatchSplit}(\boldsymbol{\hat{F}}).
\end{equation}

According to existing works \cite{fan2020pranet,Liu_2018_ECCV,fan2021concealed}, a set of convolutional layers with different kernel sizes can enlarge the receptive fields of network for improved performance.
Motived by this, we incorporate the texture enhanced modules (TEMs) \cite{fan2021concealed}, which are advanced receptive filed blocks, into our Siamese transformer by feeding the side-out features to TEMs before passing to the subsequent modules.
Mathematically, we define the final features as
\begin{equation}
    \begin{aligned}
        \boldsymbol{{F}}_{\text{r}} & =\operatorname{TEM}(\boldsymbol{\hat{F}}_{\text{r}}) \in \mathbb{R}^{H \times W \times C}, \\
        \boldsymbol{{F}}_{\text{a}} & =\operatorname{TEM}(\boldsymbol{\hat{F}}_{\text{a}}) \in \mathbb{R}^{H \times W \times C},
    \end{aligned}
\end{equation}
where $W$, $H$, and $C$ denoting the width, height, and number of channels of the feature maps, respectively. According to \cite{fan2020pranet}, we set $C$ to 32.

\begin{algorithm}[t]
    \caption{Mixture-Attention Computation.}\label{algorithm:Mixture-Attention}
    \begin{algorithmic}[1]
        \Require Input features: $\boldsymbol{F}_{\text{a}}$, $\boldsymbol{F}_{\text{r}}$
        \Ensure Refined features: $\boldsymbol{Z}_{\text{a}}, \boldsymbol{Z}_{\text{r}}$
        \State Perform feature embedding to obtain $\boldsymbol{E}_{\text{a}}$ and $\boldsymbol{E}_{\text{r}}$
        \State Concatenate $\boldsymbol{E}_{\text{a}}$ and $\boldsymbol{E}_{\text{r}}$ into $\left[\boldsymbol{E}_{\text{r}}, \boldsymbol{E}_{\text{a}}\right]$ and $\left[\boldsymbol{E}_{\text{a}}, \boldsymbol{E}_{\text{r}}\right]$
        \State Compute attention matrix $\boldsymbol{A}$ by multiplying $\left[\boldsymbol{E}_{\text{r}}, \boldsymbol{E}_{\text{a}}\right]^\top$ and $\left[\boldsymbol{E}_{\text{a}}, \boldsymbol{E}_{\text{r}}\right]$, \ie, $\boldsymbol{A} = \left[\boldsymbol{E}_{\text{r}}, \boldsymbol{E}_{\text{a}}\right]^\top \left[\boldsymbol{E}_{\text{a}}, \boldsymbol{E}_{\text{r}}\right]$
        \Comment{Eq. \eqref{equation_AequalMultiplication}}
        \State Split $\boldsymbol{A}$ into $\left[\begin{array}{cc}
                    \boldsymbol{A}_{\text{ra}} & \boldsymbol{A}_{\text{rr}} \\
                    \boldsymbol{A}_{\text{aa}} & \boldsymbol{A}_{\text{ar}}
                \end{array}\right]$ according to the meaning of sub-matrices
        \Comment{Eq. \eqref{equation_AequalMatrix}}
        \State Enhance the embedding features with normalized attention matrices:

        $\left[\begin{array}{cc}\boldsymbol{E}_{\text{r}}^{\text{(m)}}\\ \boldsymbol{E}_{\text{a}}^{\text{(s)}}\end{array}\right] = \left[\begin{array}{cc}\boldsymbol{E}_{\text{r}}\\ \boldsymbol{E}_{\text{a}}\end{array}\right] \circ
            \operatorname{softmax}
            \left(\left[\begin{array}{cc}
                    \boldsymbol{A}_{\text{ra}} \\
                    \boldsymbol{A}_{\text{aa}}
                \end{array}\right]\right),$

        $\left[\begin{array}{cc}\boldsymbol{E}_{\text{a}}^{\text{(m)}}\\ \boldsymbol{E}_{\text{r}}^{\text{(s)}}\end{array}\right] =
            \left[\begin{array}{cc}\boldsymbol{E}_{\text{a}}\\ \boldsymbol{E}_{\text{r}}\end{array}\right] \circ \operatorname{softmax}
            \left(\left[\begin{array}{cc}\boldsymbol{A}_{\text{ra}}\\ \boldsymbol{A}_{\text{rr}}\end{array}\right]\right)$

        \Comment{Eq. \eqref{equation_Softmax}}
        \State Compute $\boldsymbol{Z}_{\text{a}}$ via
        $\boldsymbol{Z}_{\text{a}} = \lambda\boldsymbol{E}_{\text{r}}^{\text{(m)}}+(\mathbf{1}-\lambda)\boldsymbol{E}_{\text{a}}^{\text{(s)}}$
        \Comment{Eq. \eqref{equation_ZequalAdd}}
        \State Compute $\boldsymbol{Z}_{\text{r}}$ via
        $\boldsymbol{Z}_{\text{r}} = \lambda\boldsymbol{E}_{\text{a}}^{\text{(m)}}+(\mathbf{1}-\lambda)\boldsymbol{E}_{\text{r}}^{\text{(s)}}$
        \Comment{Eq. \eqref{equation_ZequalAdd}}
    \end{algorithmic}
\end{algorithm}
\subsection{Mixture-Attention Module}\label{sec:MixtureAttention}
We propose a mixture-attention mechanism to capture long-range spatiotemporal relationships in videos.
It includes self-attention for intra-frame spatial relationships and mutual attention for inter-frame temporal relationships, working directly with the transformer's feature sequences.

\subsubsection{Feature Embedding}\label{sec:serialization}
The features (\ie, $\boldsymbol{F}_{\text{a}}$ and $\boldsymbol{F}_{\text{r}}$) provided by our Siamese transformer are first divided into patches, each of which is flattened and projected to an embedding.
Specifically, we denote the patch embeddings as $\boldsymbol{E}_{\text{a}}\in \mathbb{R}^{P^{2} \times NC}$ and $\boldsymbol{E}_{\text{r}} \in \mathbb{R}^{P^{2} \times NC}$, with $P$ and $N=\frac{H W}{P^{2}}$ denoting the patch size and the number of patches in each feature map.
We then add the respective position embeddings to $\boldsymbol{E}_{\text{a}}$ and $\boldsymbol{E}_{\text{r}}$ before computing the attention matrix.

\subsubsection{Attention Matrix Calculation}\label{sec:fusion}
Before computing the attention matrix, we first use concatenation operation $\left[\cdot\right]$ to combine anchor and reference feature embeddings to obtain the object embeddings $ \left[\boldsymbol{E}_{\text{a}}, \boldsymbol{E}_{\text{r}}\right] $ and
$ \left[\boldsymbol{E}_{\text{r}}, \boldsymbol{E}_{\text{a}}\right] \in \mathbb{R}^{P^{2} \times 2 N C} $.

We then calculate the attention matrix $\boldsymbol{A}$ with concatenated feature embeddings to model the long-range spatio-temporal relationships. Mathematically, $\boldsymbol{A}$ is defined as:
\begin{equation}  \label{equation_AequalMultiplication}
    \boldsymbol{A} =\left[\boldsymbol{E}_{\text{r}}, \boldsymbol{E}_{\text{a}}\right]^\top \left[\boldsymbol{E}_{\text{a}}, \boldsymbol{E}_{\text{r}}\right]
    \in \mathbb{R}^{2 N C \times 2 N C}.
\end{equation}

\subsubsection{Enhancement and Fusion}\label{sec:interaction}
We then enhance and fuse the inter-frame and intra-frame features with attention matrix $\boldsymbol{A}$, which involves two major steps.
First, we perform enhancement with the mutual-attention and self-attention sub-matrices extracted from the overall attention matrix.
Second, we fuse the enhanced feature embeddings with an addition operation.
The first step involves the matrix decomposition, where we divide the matrix $\boldsymbol{A} $ into four sub-matrices according to their actual meanings.
Mathematically, this procedure is defined as follows:
\begin{equation}  \label{equation_AequalMatrix}
    \boldsymbol{A}=\left[\begin{array}{cc}
            \boldsymbol{A}_{\text{ra}} & \boldsymbol{A}_{\text{rr}} \\
            \boldsymbol{A}_{\text{aa}} & \boldsymbol{A}_{\text{ar}}
        \end{array}\right],
\end{equation}
where $\left\{\boldsymbol{A}_{\text{ra}}, \boldsymbol{A}_{\text{rr}}, \boldsymbol{A}_{\text{aa}}, \boldsymbol{A}_{\text{ar}}\right\} \in \mathbb{R}^{N C \times N C}$ are sub-matrices with the same dimension. $\boldsymbol{A}_{\text{ra}} = \boldsymbol{A}_{\text{ar}}^{\top}$ is the mutual-attention sub-matrix.
$\boldsymbol{A}_{\text{rr}}$ and $\boldsymbol{A}_{\text{aa}}$ are the self-attention sub-matrices for $\boldsymbol{E}_{\text{r}}$ and $\boldsymbol{E}_{\text{a}}$, respectively.
From Eq. \eqref{equation_AequalMultiplication} and Eq. \eqref{equation_AequalMatrix}, it can be observed that:
\begin{equation}
    \begin{aligned}
        \boldsymbol{A}_{\text{ra}} & = \boldsymbol{A}_{\text{ar}}^{\top} = \boldsymbol{E}_{\text{r}}^\top \boldsymbol{E}_{\text{a}},
        \\
        \boldsymbol{A}_{\text{rr}} & =\boldsymbol{E}_{\text{r}}^\top \boldsymbol{E}_{\text{r}},
        \\
        \boldsymbol{A}_{\text{aa}} & =\boldsymbol{E}_{\text{a}}^\top \boldsymbol{E}_{\text{a}}.
    \end{aligned}
\end{equation}

The relationships between different sub-matrices are also illustrated \figref{fig:model_structure}.

We employ $\boldsymbol{A}_{\text{ra}}$, $\boldsymbol{A}_{\text{rr}}$, and $\boldsymbol{A}_{\text{aa}}$, to explicitly model the inter-frame temporal relationships, intra-anchor-frame spatial relationships, and intra-reference-frame spatial relationships, providing valuable spatio-temporal attention information to enhance the embedding features.

The attention matrices are normalized with $\operatorname{softmax}$ functions and then employed to enhance the embedding features. Mathematically, we define the enhanced features as:
\begin{equation}  \label{equation_Softmax}
    \begin{aligned}
        \left[\begin{array}{cc}\boldsymbol{E}_{\text{r}}^{\text{(m)}}\\ \boldsymbol{E}_{\text{a}}^{\text{(s)}}\end{array}\right] & = \left[\begin{array}{cc}\boldsymbol{E}_{\text{r}}\\ \boldsymbol{E}_{\text{a}}\end{array}\right] \circ
        \operatorname{softmax}
        \left(\left[\begin{array}{cc}
                            \boldsymbol{A}_{\text{ra}} \\
                            \boldsymbol{A}_{\text{aa}}
                        \end{array}\right]\right),
        \\
        \left[\begin{array}{cc}\boldsymbol{E}_{\text{a}}^{\text{(m)}}\\ \boldsymbol{E}_{\text{r}}^{\text{(s)}}\end{array}\right] & =
        \left[\begin{array}{cc}\boldsymbol{E}_{\text{a}}\\ \boldsymbol{E}_{\text{r}}\end{array}\right] \circ \operatorname{softmax}
        \left(\left[\begin{array}{cc}\boldsymbol{A}_{\text{ra}}\\ \boldsymbol{A}_{\text{rr}}\end{array}\right]\right),
    \end{aligned}
\end{equation}
where $\circ$ denotes the Hadamard product.
Through feature enhancement, we obtain $\boldsymbol{E}_{\text{r}}^{\text{(m)}}$ and $\boldsymbol{E}_{\text{a}}^{\text{(m)}}$ enhanced by mutual attention as well as $\boldsymbol{E}_{\text{a}}^{\text{(s)}}$ and $\boldsymbol{E}_{\text{r}}^{\text{(s)}}$ enhanced by self attention.

Before fusing the features, we split the enhanced embedding features into individual mutual- and self-attention portions and aggregate them according to their types (i.e, anchor/reference).
Mathematically, we have the fused feature embeddings $\boldsymbol{Z}_{\text{a}}\in \mathbb{R}^{P^{2} \times N C}$ and $\boldsymbol{Z}_{\text{r}} \in \mathbb{R}^{P^{2} \times N C}$ via
\begin{equation}  \label{equation_ZequalAdd}
    \begin{array}{c}
        \begin{aligned}
            \boldsymbol{Z}_{\text{a}} & = \lambda\boldsymbol{E}_{\text{r}}^{\text{(m)}}+(\mathbf{1}-\lambda)\boldsymbol{E}_{\text{a}}^{\text{(s)}}, \\
            \boldsymbol{Z}_{\text{r}} & = \lambda\boldsymbol{E}_{\text{a}}^{\text{(m)}}+(\mathbf{1}-\lambda)\boldsymbol{E}_{\text{r}}^{\text{(s)}},
        \end{aligned}
    \end{array}
\end{equation}
where the tuning factor $\lambda$ balances the contributions of mutual-attention and self-attention enhanced feature embeddings.
Please refer to Algorithm \ref{algorithm:Mixture-Attention} for the pseudo code of our mixture-attention module.

\begin{table}[t]
    \centering
    \renewcommand{\arraystretch}{1.3}
    \small
    \caption{Quantitative comparison experiments on two sub-datasets under the unseen scenario. $\uparrow$ denotes the higher, the better, and $\downarrow$ denotes the lower, the better. The best scores are in \textbf{bold}. MAST$^{\star}$ is an ablated version of our model, which is with the same backbone of PNS+ (\ie, Res2Net).}\label{tab:Quantitative}
    \vspace{5pt}
    \resizebox{0.95\textwidth}{!}{
        \begin{tabular}{r|cccccc|cccccc}
            \toprule
            \multirow{2}{*}{Model}              & \multicolumn{6}{c|}{SUN-SEG-\texttt{Easy}} & \multicolumn{6}{c}{SUN-SEG-\texttt{Hard}}                                                                                                                                                                                                         \\
                                                & $S_{\alpha}\uparrow$                       & Dice$\,\uparrow$                          & $E_{\Phi}^{mn}\uparrow$ & $F_{\beta}^{mn}\uparrow$ & $F_{\beta}^{w}\uparrow$ & Sen.$\,\uparrow$
                                                & $S_{\alpha}\uparrow$                       & Dice$\,\uparrow$                          & $E_{\Phi}^{mn}\uparrow$ & $F_{\beta}^{mn}\uparrow$ & $F_{\beta}^{w}\uparrow$ & Sen.$\,\uparrow$                                                                                                       \\
            \midrule
            UNet \cite{ronneberger2015u}       & 0.669                                      & 0.530                                     & 0.677                   & 0.528                    & 0.459                   & 0.420            & 0.670          & 0.542          & 0.679          & 0.527          & 0.457          & 0.429          \\
            UNet++ \cite{zhou2019unetplusplus} & 0.684                                      & 0.559                                     & 0.687                   & 0.553                    & 0.491                   & 0.457            & 0.685          & 0.554          & 0.697          & 0.544          & 0.480          & 0.467          \\
            SANet \cite{wei2021shallow}        & 0.720                                      & 0.649                                     & 0.745                   & 0.634                    & 0.566                   & 0.521            & 0.706          & 0.598          & 0.743          & 0.580          & 0.526          & 0.505          \\
            PraNet \cite{fan2020pranet}        & 0.733                                      & 0.621                                     & 0.753                   & 0.632                    & 0.572                   & 0.524            & 0.717          & 0.598          & 0.735          & 0.607          & 0.544          & 0.512          \\
            DCRNet \cite{yin2022duplex}        & 0.739                                      & 0.590                                     & 0.726                   & 0.658                    & 0.590                   & 0.524            & 0.732          & 0.575          & 0.713          & 0.637          & 0.573          & 0.522          \\
            LDNet \cite{zhang2022lesion}       & 0.749                                      & 0.576                                     & 0.741                   & 0.627                    & 0.557                   & 0.543            & 0.753          & 0.574          & 0.745          & 0.620          & 0.550          & 0.554          \\
            ACSNet \cite{zhang2020adaptive}    & 0.782                                      & 0.713                                     & 0.779                   & 0.688                    & 0.642                   & 0.601            & 0.783          & 0.708          & 0.787          & 0.684          & 0.636          & 0.618          \\
            UACANet \cite{kim2021uacanet}      & 0.831                                      & 0.757                                     & 0.856                   & 0.796                    & 0.754                   & 0.718            & 0.824          & 0.739          & 0.848          & 0.773          & 0.734          & 0.707          \\
            \hline
            AMD \cite{liu2021emergence}        & 0.474                                      & 0.266                                     & 0.533                   & 0.146                    & 0.133                   & 0.222            & 0.472          & 0.252          & 0.527          & 0.141          & 0.128          & 0.213          \\
            DCF \cite{zhang2021dynamic}        & 0.523                                      & 0.325                                     & 0.514                   & 0.312                    & 0.270                   & 0.340            & 0.514          & 0.317          & 0.522          & 0.303          & 0.263          & 0.364          \\
            COSNet \cite{lu2019see}            & 0.654                                      & 0.596                                     & 0.600                   & 0.496                    & 0.431                   & 0.359            & 0.670          & 0.606          & 0.627          & 0.506          & 0.443          & 0.380          \\
            PCSA \cite{gu2020pyramid}          & 0.680                                      & 0.592                                     & 0.660                   & 0.519                    & 0.451                   & 0.398            & 0.682          & 0.584          & 0.660          & 0.510          & 0.442          & 0.415          \\
            FSNet \cite{ji2021full}            & 0.725                                      & 0.702                                     & 0.695                   & 0.630                    & 0.551                   & 0.493            & 0.724          & 0.699          & 0.694          & 0.611          & 0.541          & 0.491          \\
            PNSNet \cite{ji2021progressively}  & 0.767                                      & 0.676                                     & 0.744                   & 0.664                    & 0.616                   & 0.574            & 0.767          & 0.675          & 0.755          & 0.656          & 0.609          & 0.579          \\
            MAT \cite{zhou2020matnet}          & 0.770                                      & 0.710                                     & 0.737                   & 0.641                    & 0.575                   & 0.542            & 0.785          & 0.712          & 0.755          & 0.645          & 0.578          & 0.579          \\
            SSTAN \cite{zhao2022semi}          & 0.774                                      & 0.642                                     & 0.784                   & 0.694                    & 0.634                   & 0.592            & 0.784          & 0.662          & 0.815          & 0.707          & 0.647          & 0.624          \\
            2/3D \cite{puyal2020endoscopic}    & 0.786                                      & 0.722                                     & 0.777                   & 0.708                    & 0.652                   & 0.603            & 0.786          & 0.706          & 0.775          & 0.688          & 0.634          & 0.607          \\
            PNS+ \cite{ji2022video}            & 0.806                                      & 0.756                                     & 0.798                   & 0.730                    & 0.676                   & 0.630            & 0.797          & 0.737          & 0.793          & 0.709          & 0.653          & 0.623          \\
            \hline
            MAST$^{\star}$                      & 0.830                                      & 0.771                                     & 0.839                   & 0.762                    & 0.720                   & 0.698            & 0.848          & 0.781          & 0.874          & 0.776          & 0.738          & 0.754          \\
            \textbf{\ourmodel}                  & \textbf{0.845}                             & \textbf{0.784}                            & \textbf{0.898}          & \textbf{0.819}           & \textbf{0.770}          & \textbf{0.755}   & \textbf{0.861} & \textbf{0.803} & \textbf{0.914} & \textbf{0.816} & \textbf{0.777} & \textbf{0.811} \\
            \bottomrule
        \end{tabular}}
    \label{tab:without-booktabs}
\end{table}

\subsection{Parallel Decoders}\label{sec:ParallelDecoders}
The feature embeddings $\boldsymbol{Z}_{\text{a}}$ and $\boldsymbol{Z}_{\text{r}}$ are rearranged into feature maps $\boldsymbol{Z}_{\text{a}}^{'}\in \mathbb{R}^{H \times W \times C}$ and $\boldsymbol{Z}_{\text{r}}^{'} \in \mathbb{R}^{H \times W \times C}$, which, together with other levels of features from different encoder layers, are fed to two parallel decoders.
Each decoder includes three layers, each of which consists of Neighbor Connection Decoder (NCD) and multiple Group-Reversal Attention (GRA) blocks \cite{fan2021concealed}.
Interested readers can refer to \cite{fan2021concealed} for the architecture of each decoder.
Thanks to the reverse-guidance and group-guidance operations, the decoder can gradually refine the rough prediction by different feature pyramids and exploit the multi-level features to predict the corresponding segmentation probability maps, \ie, $\{\hat{\boldsymbol{Y}}^{(i)}_{\text{a}}\}^{4}_{i=1}$ and $\{\hat{\boldsymbol{Y}}^{(i)}_{\text{r}}\}^{4}_{i=1}$ for the anchor and reference frames, respectively. It is worth noting that we adopt deep supervision to supervise each level of the decoder, which, therefore, results in four output segmentation maps for each frame.

\subsection{Loss Function}\label{sec:LossFunction}
We use a hybrid loss function to train our \ourmodel{}, which combines the Binary Cross-Entropy (BCE) loss and Intersection over Union (IoU) loss \cite{zhao2019egnet}.
The total loss function is defined as
\begin{equation}
    \mathcal{L} = \sum_{i = 1}^{4} \sum_{j \in \{\text{a}, \text{r}\}} \mathcal{L}_{\text{BCE}}^{w}\left(\boldsymbol{Y}, \hat{\boldsymbol{Y}}^{(i)}_{j}\right)+\mathcal{L}_{\text{IoU}}^{w}\left(\boldsymbol{Y}, \hat{\boldsymbol{Y}}^{(i)}_{j}\right),
\end{equation}
where $\boldsymbol{Y}_{\text{a}}$ and $\boldsymbol{Y}_{\text{r}}$ represent the ground truth segmentation maps for the anchor and reference frames, respectively.
$\mathcal{L}_{\text{BCE}}^{w}(\cdot)$ denotes the weighted BCE loss, which assigns a weight to each pixel based on the difference between the center pixel of feature map and its surroundings to better constrain the model to focus on the hard pixels of the target.
$\mathcal{L}_{\text{IoU}}^{w}(\cdot)$ denotes the weighted IoU loss, which adds pixel weights to the normal IoU loss to constrain the global region differently.

\section{Experiments} \label{sec:Experiments}
In this section, we will present the experimental details, including the dataset and training settings (see Section \secref{sec:ImplementationDetails}), the quantitative (see Section \secref{sec:QuantitativeResults}) and qualitative (see Section \secref{sce:QualitativeResults}) results, and the the ablation study (see Section \secref{sec:Ablation Study}).

\subsection{Implementation Details} \label{sec:ImplementationDetails}
\subsubsection{Datasets}
In our experiments, we use the largest-scale VPS benchmark to date, the SUN-SEG dataset. This dataset is created by re-organizing the colonoscopy video database from Showa University and Nagoya University. It contains 1,106 short video clips with a total of 158,690 frames, including 378 positive and 728 negative cases. We follow the same training/testing setting as in PNS+ \cite{ji2022video} and only conduct experiments on positive cases. For training, we use 40\% of the SUN-SEG dataset, including 112 clips with 19,544 frames. For testing, we use two unseen testing subsets, namely SUN-SEG-\texttt{Easy} with 54 clips (12,522 frames) and SUN-SEG-\texttt{Hard} with 119 clips (17,070 frames).

\subsubsection{Training Details}
Our MAST model is trained on a NVIDIA 3060 GPU.
Before training, we load the ImageNet pre-training weights for PVTv2-B2 \cite{wang2022pvt} and adjust the input images to $352\times352$.
During training, the initial learning rate is set to \num{1e-5}, and the learning rate decays by a specific ratio for every ten training epochs.
Meanwhile, the number of epochs is set to $30$, and the batch size is set to $24$.
Each batch consists of a reference frame at timestamp $t=T$ and an anchor frame at $t=T-\Delta$, where the time interval $\Delta=2$.
Additionally, we set the attention weighting factor $\lambda$ to $0.7$ through parameter grid search.

\subsubsection{Evaluation Metrics}
To conduct a comprehensive analysis of the experiments, we employ the following six metrics for evaluating the results.
Assuming the ground truth map is $\boldsymbol{Y}$ and the binary prediction map is $\hat{\boldsymbol{Y}}$.

(a) \textbf{Dice Coefficient (Dice).} The Dice coefficient is a statistical measure used to assess the similarity between two sets. In the realm of image segmentation, it serves as a quantification of the degree of overlap between the predicted segmentation and the ground truth segmentation. Mathematically, it is defined as:
\begin{equation}
    \text{Dice}=\frac{2 \lvert \hat{\boldsymbol{Y}} \cap \boldsymbol{Y} \rvert}{ \lvert \hat{\boldsymbol{Y}} \rvert + \lvert \boldsymbol{Y} \rvert},
\end{equation}
where $\lvert \hat{\boldsymbol{Y}} \rvert + \lvert \boldsymbol{Y} \rvert$ denotes the sum of pixels in two sets ($\hat{\boldsymbol{Y}}$ and $\boldsymbol{Y}$).

(b) \textbf{F-measure ($F_{\beta}^{mn}$).} The F-measure is a harmonic mean of precision and recall, with a weighting factor $\beta$. It offers a more comprehensive evaluation of the segmentation results and can be calculated as follows:
\begin{equation}
    F_{\beta}=\frac{(1+\beta^2) \text{Precision} \times \text{Recall}}{\beta^2 \times \text{Precision}+\text{Recall}},
\end{equation}
where $\text{Precision}=\frac{\lvert \hat{\boldsymbol{Y}} \cap \boldsymbol{Y} \rvert}{\lvert \hat{\boldsymbol{Y}} \rvert}$ and $\text{Recall}=\frac{\lvert \hat{\boldsymbol{Y}} \cap \boldsymbol{Y} \rvert}{\lvert \boldsymbol{Y} \rvert}$.
By employing an adaptive threshold to binarize the original prediction map and subsequently calculating the average of the results ($F_{\beta}$), we can derive the mean F-measure ($F_{\beta}^{mn}$). The adaptive threshold is precisely defined as twice the average pixel intensity of the original prediction map $\hat{\boldsymbol{Y}}_o$:
\begin{equation}
    \text{AdaptiveThreshold}=\frac{2}{w \times h} \sum^w_{x=1} \sum^h_{y=1} \hat{\boldsymbol{Y}}_o(x,y),
\end{equation}
where $h$ and $w$ denote the height and the width of the map, respectively.

(c) \textbf{Weighted F-measure ($F_{\beta}^{w}$).} Margolin \etal \cite{2014Margolin} introduced the concepts of weighted Precision ($\text{Precision}^w$) and weighted Recall ($\text{Recall}^w$), rectifying three erroneous assumptions regarding interpolation, dependency, and equal-importance. Building upon these advancements, the F-measure is extended to a weighted F-measure, thereby providing a comprehensive approach for evaluating both non-binary and binary maps. $F_{\beta}^{w}$ is defined as:

\begin{equation}
    F_{\beta}^{w}=\frac{(1+\beta^2) \text{Precision}^w \times \text{Recall}^w}{\beta^2 \times \text{Precision}^w+\text{Recall}^w},
\end{equation}
where $\beta$ represents the detection efficacy in relation to a user who assigns $\beta$ times more significance to $\text{Recall}^w$ compared to $\text{Precision}^w$.

(d) \textbf{Sensitivity (Sen.).} Sensitivity, which is employed to assess the proportion of accurately predicted positive instances out of all segmentation results, can be defined as follows:
\begin{equation}
    \text{Sen.}=\frac{\lvert \hat{\boldsymbol{Y}} \cap \boldsymbol{Y} \rvert}{\lvert \boldsymbol{Y} \rvert}.
\end{equation}

All results of Sensitivity in this paper are also the mean values.

(e) \textbf{Structure measure ($S_{\alpha}$) \cite{fan2017structure}.}
The structure measure simultaneously evaluates region-aware and object-aware structural similarity between a target and a ground-truth map. For binary maps, region-awareness emphasizes luminance, contrast, and dispersion probability comparisons. Mathematically, it can be expressed as follows:
\begin{equation}
    S_{\alpha}=\alpha \times S_o(\hat{\boldsymbol{Y}},\boldsymbol{Y})+(1-\alpha) \times S_a(\hat{\boldsymbol{Y}},\boldsymbol{Y}),
\end{equation}
where the $\alpha$ denotes the weighting factor, set to 0.5 by default.

(f) \textbf{Enhanced-alignment measure (E-measure) \cite{fan2021cognitive}.} The E-measure, designed specifically for binary map evaluation, effectively combines image-level statistics and local pixel matching information. Its definition is as follows:
\begin{equation}
    E_{\Phi}=\frac{1}{w \times h} \sum^w_{x=1} \sum^h_{y=1} \phi \left(\hat{\boldsymbol{Y}}(x,y),\boldsymbol{Y}(x,y)\right),
\end{equation}
where $\phi$ represents the enhanced alignment matrix, while $h$ and $w$ denote the map's height and width, respectively. Table \ref{tab:Quantitative} shows the mean values of E-measure ($E_{\Phi}^{mn}$).

The metrics employed encompass structural similarity, intersectional similarity, precision, and recall, all of which are instrumental in evaluating the accuracy of the model's predictions and their concordance with the ground truth.

\begin{figure}[htbp]
    \centering
    \includegraphics[trim={1.5cm 2cm 1.5cm 2cm},clip,width=\linewidth]{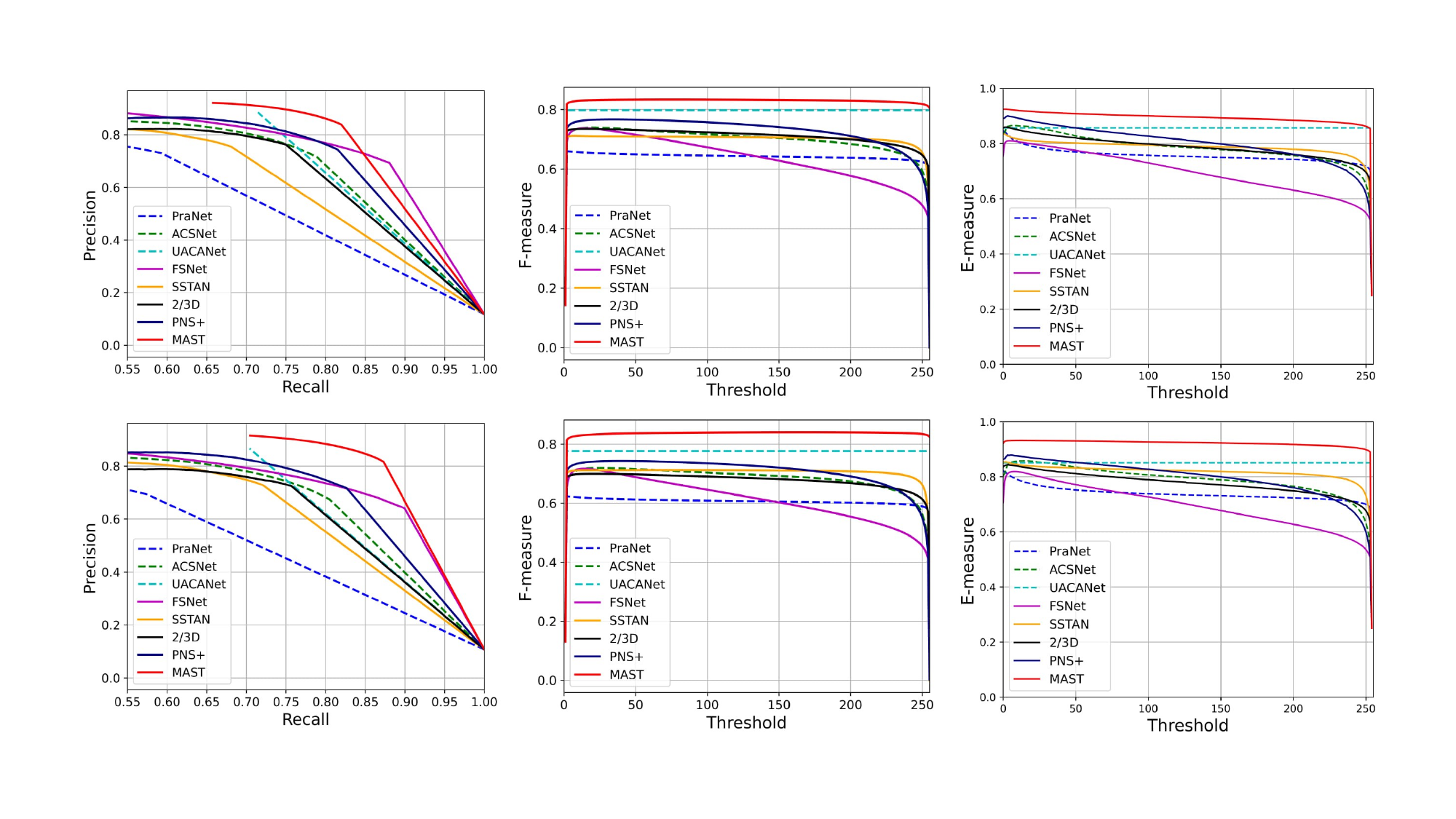}
    \caption{Comparison of Precision-Recall, F-measure, and E-measure curves between cutting-edge competitors and our \ourmodel{} on the SUN-SEG-\texttt{Easy} (1$^{st}$ row) and SUN-SEG-\texttt{Hard} datasets (2$^{nd}$ row).}
    \label{fig:Curve}
\end{figure}

\subsection{Quantitative Results} \label{sec:QuantitativeResults}
Table \ref{tab:Quantitative} presents a quantitative comparison among our models (MAST$^{\star}$ and MAST) and several state-of-the-art models, encompassing eight image-based models and ten video-based models. To ensure a fair and rigorous comparison, we employ the SUN-SEG dataset for training across all models while maintaining default settings for optimal results. Under these standardized conditions, we conducted VPS benchmark \cite{ji2022video} tests on both the SUN-SEG-\texttt{Easy} and SUN-SEG-\texttt{Hard} datasets.

The experimental results reveal noteworthy distinctions in the performance of the image-based model, UACANet, when compared to the video-based model, PNS+. This disparity is evident across both test sets. For instance, when examining the evaluation metrics for structural accuracy within the SUN-SEG-\texttt{Easy} and SUN-SEG-\texttt{Hard} datasets, UACANet exhibits superior performance with results of 0.831 and 0.824, as opposed to PNS+, which achieves results of 0.806 and 0.797, respectively. Similarly, in the context of the E-measure evaluation metrics for these two test sets, UACANet consistently outperforms PNS+ with results of 0.856 and 0.848, while PNS+ achieves results of 0.798 and 0.793. These observations underscore UACANet's heightened efficacy in the domain of image-based segmentation.
Founded on a CNN framework, UACANet augments its performance by incorporating contextual features of uncertain regions, thereby enhancing its ability to discern boundary information. In contrast, PNS+ adopts a transformer-based architecture that leverages attention mechanisms to capture global-to-local information from video frames. The discernible difference in model performance can be attributed to the inherent sensitivity of CNNs to localized image information, enabling UACANet to meticulously extract boundary cues and consequently achieve superior segmentation results. On the other hand, PNS+ excels in learning long-range inter-frame dependencies, which proves advantageous in tracking target movement within video sequences.

\ourmodel{}, amalgamates the strengths of transformers and CNNs to enhance the localization, tracking, and segmentation of polyp targets within video data. Siamese transformer network effectively learns paired frames in video streams. The integration of inter-frame dependencies and intra-frame features is facilitated by the Mixture-Attention module, while a coarse location map guides the continuous refinement of segmentation accuracy. The synergy of these components empowers MAST to execute its tasks with exceptional precision. Furthermore, our exploration of an alternative architecture, wherein Res2Net replaces the Siamese transformer as the model backbone, yields noteworthy insights. The results, presented in the penultimate row of Table \ref{tab:Quantitative}, underscore the remarkable performance of MAST$^{\star}$, surpassing that of PNS+. This underscores the remarkable efficacy of our meticulously designed architecture.

\figref{fig:Curve} illustrates the Precision-Recall (PR), F-measure, and E-measure curves for both the SUN-SEG-\texttt{Easy} and SUN-SEG-\texttt{Hard} datasets, offering a visual assessment of the results presented in Table \ref{tab:Quantitative}. Notably, our proposed method, \ourmodel{}, exhibits superior performance across six representative images. This superiority is evident through the MAST-generated Precision-Recall curves, which encompass the largest area under the curve, and the F-measure and E-measure curves that demonstrate the highest degree of similarity. These results unequivocally validate the effectiveness and inherent advantages of \ourmodel{}.

\begin{figure}[htbp]
    \centering
    {\includegraphics[trim={2.0cm 0.5cm 2.0cm 0cm},width=1\linewidth]{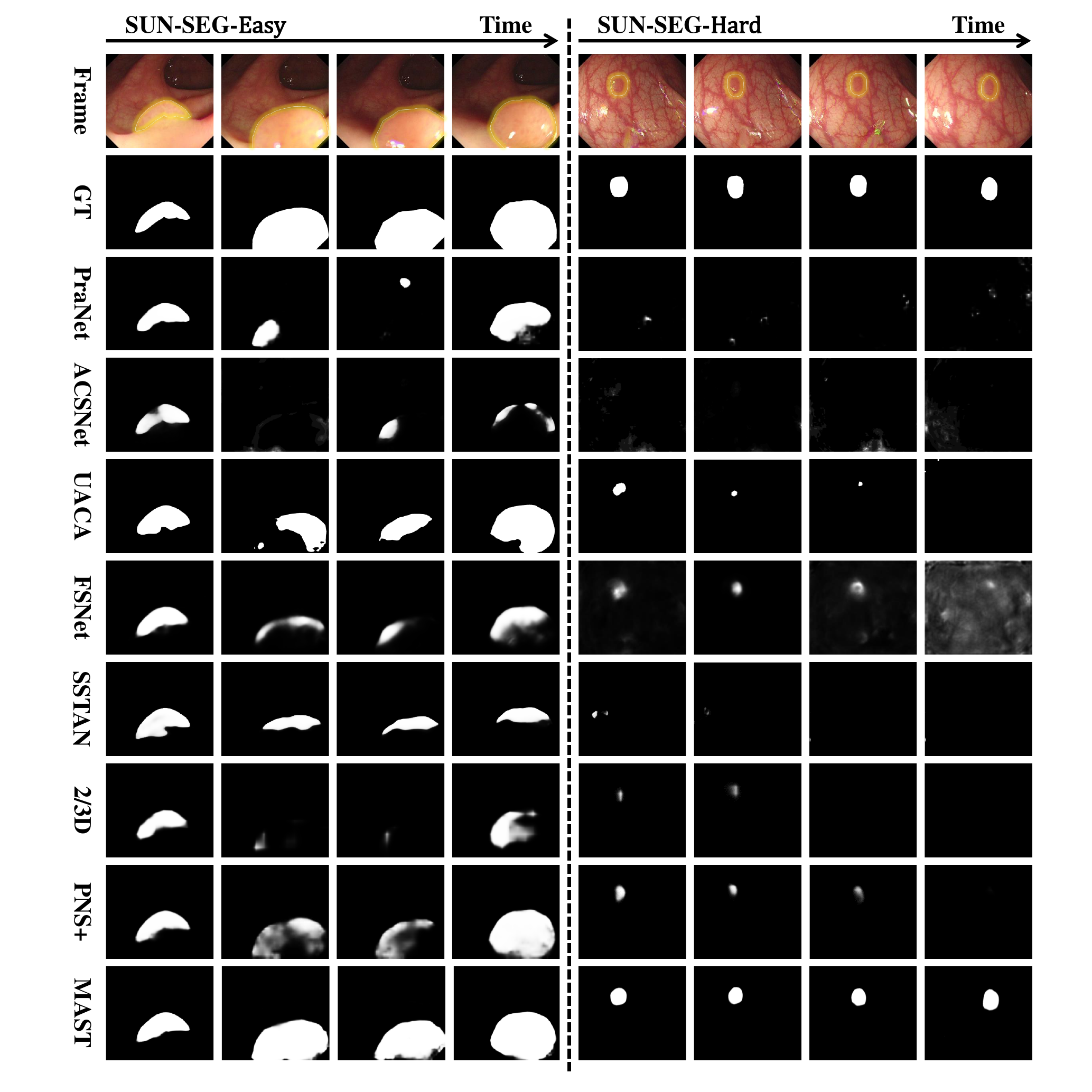}}
    \vspace{-20pt}
    \caption{
        The qualitative comparison of cutting-edge competitors and our \ourmodel.
    }
    \label{fig:Qualitative}
\end{figure}

\subsection{Qualitative Results} \label{sce:QualitativeResults}

\figref{fig:Qualitative} shows the segmentation results obtained by multiple models, including \ourmodel{}, across consecutive video frames. The visual depiction in figure underscores that \ourmodel{} consistently achieves a high degree of precision in segmenting polyps resembling colonic mucosa, closely aligning with the ground truth. Conversely, other methodologies employed in this study fail to achieve comparable segmentation performance during certain temporal intervals.

In our comparative analysis, we observe a marked disparity in the performance of competing models when applied to distinct subsets of the dataset. Specifically, these models exhibit a heightened sensitivity to conspicuous targets within the SUN-SEG-\texttt{Easy} dataset, while struggling to discern and track smaller, more challenging targets within the SUN-SEG-\texttt{Hard} dataset. Conversely, our proposed model excels in the recognition of diminutive polyps and demonstrates robust tracking capabilities for challenging targets within video sequences. These accomplishments can be attributed to the effective utilization of spatiotemporal cues, harnessed through the integration of our Siamese transformer and mixture-attention module.

\subsection{Ablation Study} \label{sec:Ablation Study}
To establish the robustness and efficacy of our core design, we conducted ablation experiments targeting pivotal components and critical parameters within our model architecture. These elements encompassed the Siamese transformer module, the Mixture-Attention module, as well as the parameters denoted as $\lambda$ and $\Delta$. It is worth noting that all configurations employed in the ablation experiments adhered to the specifications outlined in Section \secref{sec:ImplementationDetails}. In the interest of clarity and meaningful comparison, our evaluation process was centered on three specific metrics: structure measure ($S_{\alpha}$), Dice Coefficient (Dice), and sensitivity (Sen.). The results of the ablation experiments are shown in Table \ref{tab:Components}, Table \ref{tab:lambda}, and Table \ref{tab:TimeInterval}.

\begin{table}[t]
    \centering
    \scriptsize
    \renewcommand{\arraystretch}{1.3}
    \renewcommand{\tabcolsep}{2mm}
    \caption{Ablation study for two core modules of \ourmodel. The ``M-A'' in table means the Mixture-Attention module.}\label{tab:Components}
    \begin{tabular}{ccc|ccc|ccc}
        \toprule
        \multirow{2}{*} {Baseline} & \multirow{2}{*} {Siamese} & \multirow{2}{*} {M-A} & \multicolumn{3}{c|}{SUN-SEG-\texttt{Easy}} & \multicolumn{3}{c}{SUN-SEG-\texttt{Hard}}                                                                                 \\
                                   &                           &                       & $S_{\alpha}\uparrow$                       & Dice$\,\uparrow$                          & Sen.$\,\uparrow$ & $S_{\alpha}\uparrow$ & Dice$\,\uparrow$ & Sen.$\,\uparrow$ \\
        \midrule
        \checkmark                 &                           &                       & 0.805                                      & 0.725                                     & 0.688            & 0.824                & 0.746            & 0.727            \\
        \checkmark                 & \checkmark                &                       & 0.825                                      & 0.757                                     & 0.716            & 0.844                & 0.775            & 0.761            \\
        \checkmark                 &                           & \checkmark            & 0.831                                      & 0.767                                     & 0.731            & 0.852                & 0.789            & 0.779            \\
        \checkmark                 & \checkmark                & \checkmark            & \textbf{0.845}                             & \textbf{0.784}                            & \textbf{0.755}   & \textbf{0.861}       & \textbf{0.803}   & \textbf{0.811}   \\
        \bottomrule
    \end{tabular}
\end{table}

\subsubsection{Effectiveness of Core Modules}
In this section, we explore the contribution of the Siamese transformer module and the Mixture-Attention module to \ourmodel{}. In Table \ref{tab:Components}, we present our baseline model (1$^{st}$ row), which employs a PVTv2-B2 backbone and a pair of CNN-based decoders.

\myPara{Effectiveness of Siamese Transformer Module.}
We introduce the Siamese transformer module into the baseline model to assess its impact on \ourmodel{}. The results of this incorporation are presented in the 2$^{nd}$ row of Table \ref{tab:Components}. In comparison to the baseline, this variant model exhibits notable improvements across both test datasets. The most substantial enhancement in evaluation metrics is observed in the case of the SUN-SEG-\texttt{Easy} dataset, with the Dice rising from 0.725 to 0.757. Similarly, on the SUN-SEG-\texttt{Hard} dataset, the Sensitivity metric increases from 0.727 to 0.761. These results substantiate the efficacy of the Siamese network in enhancing the model's capacity to acquire comprehensive features of polyps. Furthermore, it facilitates the model's ability to focus on complementary attributes within diverse input data, thereby enhancing its capacity to extract target information.

\myPara{Effectiveness of Mixture-Attention Module.}
Our comprehensive model, as presented in the 4$^{th}$ row, combines both the aforementioned modules, leading to further performance enhancements, particularly on the challenging SUN-SEG-\texttt{Hard} dataset. These results underscore the synergistic impact achieved through the concurrent utilization of these modules.

\begin{table}[t]
    \centering
    \scriptsize
    \renewcommand{\arraystretch}{1.3}
    \caption{Ablation study for the different weighting factors $\lambda$.}    \label{tab:lambda}
    \begin{tabular}{c|ccc|ccc}
        \toprule
        \multirow{2}{*} {Weighting Factor} & \multicolumn{3}{c|}{SUN-SEG-\texttt{Easy}} & \multicolumn{3}{c}{SUN-SEG-\texttt{Hard}}                                                                       \\
                                           & $S_{\alpha}\uparrow$                       & Dice$\,\uparrow$                          & Sen.$\,\uparrow$
                                           & $S_{\alpha}\uparrow$                       & Dice$\,\uparrow$                          & Sen.$\,\uparrow$                                                    \\
        \midrule
        $\lambda=0$                        & 0.826                                      & 0.761                                     & 0.716            & 0.845          & 0.784          & 0.761          \\
        $\lambda=0.3$                      & 0.834                                      & 0.773                                     & 0.727            & 0.848          & 0.789          & 0.769          \\
        $\lambda=0.5$                      & 0.838                                      & 0.775                                     & 0.746            & 0.851          & 0.790          & 0.782          \\
        $\lambda=0.7$                      & \textbf{0.845}                             & \textbf{0.784}                            & \textbf{0.755}   & \textbf{0.861} & \textbf{0.803} & \textbf{0.811} \\
        $\lambda=1$                        & 0.832                                      & 0.765                                     & 0.721            & 0.845          & 0.776          & 0.761          \\
        \bottomrule
    \end{tabular}
\end{table}

\subsubsection{Lambda Setting}
To ascertain the optimal fusion coefficient $\lambda$ as defined in Eq. \eqref{equation_ZequalAdd}, we conducted an empirical examination of its impact on the mixture-attention module, as summarized in Table 2. A range of $\lambda$ values, was employed in this experimental investigation, \ie, $\lambda=\{0,0.3, 0.5, 0.7, 1\}$.
The results of our investigation reveal that the model configuration with $\lambda=0.7$ yields the most favorable performance compared to other values. Upon substituting "lambda=0.7" back into Eq. \eqref{equation_ZequalAdd}, we derive the ensuing equation:

\begin{equation}  \label{equation_lambda_0.7}
    \begin{array}{c}
        \begin{aligned}
            \boldsymbol{Z}_{\text{a}} & = 0.7\boldsymbol{E}_{\text{r}}^{\text{(m)}}+0.3\boldsymbol{E}_{\text{a}}^{\text{(s)}}, \\
            \boldsymbol{Z}_{\text{r}} & = 0.7\boldsymbol{E}_{\text{a}}^{\text{(m)}}+0.3\boldsymbol{E}_{\text{r}}^{\text{(s)}}.
        \end{aligned}
    \end{array}
\end{equation}

The optimal model segmentation outcomes are achieved at higher proportions of the mutual attention matrix, as indicated by Eq. \eqref{equation_lambda_0.7}. This underscores the model's reliance on long-distance spatiotemporal relationships among video frames for polyp segmentation. Consequently, we infer that the Mixture-Attention module's key function is to capture temporal information within the video stream and integrate distant spatiotemporal features with intra-frame characteristics, enabling precise target motion tracking in relation to frame data.
\begin{table}[t]
    \centering
    \scriptsize
    \renewcommand{\arraystretch}{1.3}
    \renewcommand{\tabcolsep}{2.4mm}
    \caption{Ablation study for different time intervals $\Delta$ of frame-taking strategy.}    \label{tab:TimeInterval}
    \begin{tabular}{c|ccc|ccc}
        \toprule
        \multirow{2}{*} {Time Interval} & \multicolumn{3}{c|}{SUN-SEG-\texttt{Easy}} & \multicolumn{3}{c}{SUN-SEG-\texttt{Hard}}                                                                       \\
                                        & $S_{\alpha}\uparrow$                       & Dice$\,\uparrow$                          & Sen.$\,\uparrow$
                                        & $S_{\alpha}\uparrow$                       & Dice$\,\uparrow$                          & Sen.$\,\uparrow$                                                    \\
        \midrule
        $\Delta$=1                      & 0.841                                      & 0.778                                     & 0.747            & 0.853          & 0.791          & 0.786          \\
        $\Delta$=2                      & \textbf{0.845}                             & \textbf{0.784}                            & \textbf{0.755}   & \textbf{0.861} & \textbf{0.803} & \textbf{0.811} \\
        $\Delta$=3                      & 0.831                                      & 0.779                                     & 0.740            & 0.852          & 0.789          & 0.782          \\
        $\Delta$=5                      & 0.831                                      & 0.771                                     & 0.731            & 0.849          & 0.781          & 0.779          \\
        \bottomrule
    \end{tabular}
\end{table}
\subsubsection{Time Interval Setting}
We conducted experiments to determine the optimal time interval for MAST to learn inter-frame temporal dependencies, with four strategies ($\Delta=\{1,2,3,5\}$). Table \ref{tab:TimeInterval} presents the experimental outcomes. The most favorable results occur with $\Delta=2$. The results indicate that both small and large time intervals hinder the model's ability to capture spatiotemporal dependencies. A small interval results in high feature repetition and limited spatiotemporal learning, while a large interval weakens frame connections, leading to diminished performance as $\Delta$ exceeds $2$.

\begin{table}[t]
    \centering
    \scriptsize
    \renewcommand{\arraystretch}{1.3}
    \renewcommand{\tabcolsep}{0.8mm}
    \caption{Comparison of computational complexity and model size of top-tier competitors and our \ourmodel{}.}    \label{tab:flops&params}
    \begin{tabular}{c|ccccccc}
        \toprule
        Metric     & PraNet & ACSNet & UACANet & FSNet & MAT    & PNS+  & {MAST} \\
        \midrule
        FLOPs (G)  & 13.15  & 21.88  & 17.41   & 35.33 & 83.01  & 53.24 & 21.02  \\
        Params (M) & 30.50  & 29.45  & 24.86   & 83.42 & 119.24 & 9.79  & 25.69  \\
        \bottomrule
    \end{tabular}
\end{table}

\subsubsection{Parameters and Flops}
We conducted an extensive experiments of the computational complexity and model size associated with various models. The results are presented in Table \ref{tab:flops&params}, where ``FLOPs'' denotes computational complexity, and ``Params'' denotes model size. The results reveal that our model effectively achieves a favorable equilibrium between computational efficiency and parameter dimensions when compared to SOTA models. This observation, when considered alongside the empirical findings delineated in Table \ref{tab:Quantitative}, substantiates the significant performance enhancement of our model, even when operating under conditions characterized by nearly identical parameter and FLOPs scales.

\section{Conclusion} \label{sec:Conclusion}
In this paper, we propose \ourmodel, a novel video polyp segmentation network based on a Siamese transformer and a mixture-attention mechanism.
Our network effectively models spatiotemporal relationships, enhancing feature learning for accurate polyp segmentation.
We evaluate our model on a large-scale benchmark dataset SUN-SEG. The results demonstrate that our model outperforms SOTA methods, both quantitatively and qualitatively. Further ablation experiments validate the effectiveness of our proposed components.
Future work will focus on extending our model to more challenging medical video segmentation tasks.

\bibliographystyle{elsarticle-harv}
\bibliography{refs}
\end{document}